\documentclass{article}

\usepackage{arxiv}

\usepackage[utf8]{inputenc} 
\usepackage[T1]{fontenc}    
\usepackage{hyperref}       
\usepackage{url}            
\usepackage{booktabs}       
\usepackage{amsfonts}       
\usepackage{nicefrac}       
\usepackage{microtype}      
\usepackage{lipsum}
\usepackage{graphicx}

\newcommand{\xv}{\mathbf{x}}

\newcommand{\Wv}{\mathbf{W}}
\newcommand{\bv}{\mathbf{b}}
\newcommand{\hv}{\mathbf{h}}
\newcommand{\zv}{\mathbf{z}}
\newcommand{\sv}{\mathbf{s}}
\newcommand{\xt}{\mathbf{\tilde{x}}}
\newcommand{\xh}{\mathbf{\hat{x}}}

\newcommand{\hh}{\mathbf{\hat{h}}}

\usepackage[swedish]{babel}

\usepackage{amsmath,amsthm,verbatim,amssymb,amsfonts,amscd}

\usepackage[vlined, figure]{algorithm2e}  

\usepackage{color}
\usepackage{float}
\usepackage[tight]{subfigure}
\usepackage{pgf}
\usepackage{enumitem}
\usepackage{comment}
\usepackage{caption}

\title{Establishing strong imputation performance of a denoising autoencoder in a wide range of missing data problems}

\author{Najmeh Abiri\thanks{Department of Astronomy and Theoretical Physics, Lund University, Lund, Sweden} \\ najmeh@thep.lu.se \and Bj\"orn Linse\footnotemark[1] \\bjorn.linse@thep.lu.se \and Patrik Ed\'en\footnotemark[1]\\patrik@thep.lu.se \and Mattias Ohlsson\thanks{Center for Applied Intelligent Systems Research, Halmstad University, Halmstad, Sweden}     \footnotemark[1] \\mattias@thep.lu.se}

\begin{document}
\maketitle

\begin{abstract}
Dealing with missing data in data analysis is inevitable.  Although powerful imputation methods that address this problem exist, there is still much room for improvement. In this study, we examined single imputation based on deep autoencoders, motivated by the apparent success of deep learning to efficiently extract useful dataset features. We have developed a consistent framework for both training and imputation. Moreover, we benchmarked the results against state-of-the-art imputation methods on different data sizes and characteristics. The work was not limited to the one-type variable dataset; we also imputed missing data with multi-type variables, e.g., a combination of binary, categorical, and continuous attributes. To evaluate the imputation methods, we randomly corrupted the complete data, with varying degrees of corruption, and then compared the imputed and original values. In all experiments, the developed autoencoder obtained the smallest error for all ranges of initial data corruption. 
\end{abstract}

\section{Introduction}

The presence of missing data is a practical challenge for many researchers working in data mining. Many techniques have been proposed to impute (estimate) missing data \cite{Little:2002}. The most straightforward approach, complete-case analysis, disregards samples with missing values. In many cases, such as medical data, this means a severe reduction of the data size, besides the obvious problem of introducing biases. Simple mean substitution is a common approach but can also introduce bias and completely ignores any correlation among the predictor variables. An alternative is to use nearest neighbor methods, where observations with the smaller distance to the missing data are used in the imputation process~\cite{troyanskaya_knn}. Here one can find weighted nearest neighbor~\cite{knn} approaches with good performances. Both mean imputation and nearest neighbor methods are single imputation techniques, whereas multiple imputations have the advantage to create a probability distribution of imputation values that consider uncertainty in the estimation and are generally regarded as the best approach to missing data imputation~\cite{mice}. However, multiple imputations may have problems in converging to a reasonable stationary distribution~\cite{raghunathan2001multivariate}.

Along with the developments and success of deep learning, there has been an increasing interest in autoencoders~\cite{Hinton:2006} as a tool for manifold learning and finding good representations of data. An example of efficient usage of such autoencoder representations can be found for the task of human pose recovery. Hong et al. \cite{Hong:2015} used an autoencoder architecture to reconstruct 3D poses from 2D silhouettes. Further algorithmic developments have introduced the denoising autoencoder~\cite{Vincent:2008} and the variational autoencoder~\cite{Kingma:2013}, where the latter extends the autoencoder to a generative model. Although the denoising method was mainly introduced to extract \textit{good and useful} features from the input data; it naturally extends to imputation of missing data. 
Different forms of autoencoders have been proposed in connection with the restoration of sensor data~\cite{Narayanan:2002, Thompson:2003}. There are also methods combining autoencoder training with genetic algorithms in missing data applications~\cite{Marivate:2007, Mohamed:2007, Nelwamondo:2007}, mostly using shallow autoencoder architectures. Deep learning approaches can be found in the analysis of traffic data imputation~\cite{Duan:2014} and for dealing with missing data in electronic health records~\cite{Jones:2016} and in genomic data \cite{Qiu:2018}. Using autoencoders for multiple imputation has also been proposed \cite{Gondara:2018}, although for a true generative model approach the variational autoencoder is a more natural choice.

In this paper, we study the use of denoising autoencoders for single imputation of missing data. We present a consistent framework aimed for general tabular datasets. The proposed architecture follows a butterfly construction with mirrored decoder weighs and pre-training of encoder layers. To handle missing data during model training, the essential modification of the original denoising autoencoder is to mask all unknown values from the loss function. Furthermore, we allow for imputation of data with mixed variable types using a sum of appropriate (masked) loss functions. We have tested the framework on tabular datasets with different characteristics (including image data) and compared the results against other popular and imputation methods. For all numerical tests, using a broad range of initial missing values on each of the datasets, the proposed autoencoder framework obtained the smallest errors. Unlike other imputation methods, the autoencoder network can be exported and applied on similar data after the usual model selection and training procedure.

Section~\ref{sec:methods} contains a brief introduction to stacked denoising autoencoders and gives a detailed description of the proposed imputation method. In Sections~\ref{sec:datasets}~\&~\ref{sec:experiments} there is a description of the datasets used and details of how the experiments were carried out. Section~\ref{sec:results} presents the results and a discussion and conclusion can be found in Section \ref{sec:discussion}.

\section{Methods}\label{sec:methods}
\subsection{Autoencoder structure}
Autoencoders are unsupervised neural networks consisting of an encoder and a decoder part. The encoder is trained to encode input data to a typically lower dimensional representation (one way to avoid identity mappings). Then the decoder is trained to reconstruct the input data from this representation. Such networks can capture the useful features~\cite{chapelle}, which makes them suitable for imputation of missing data. 

\begin{figure}
\centering
\includegraphics[scale=0.7]{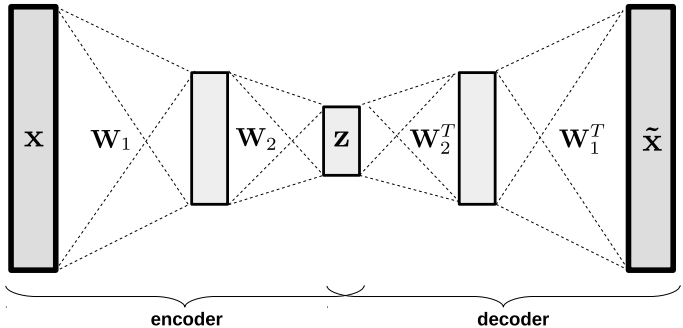}
\caption{The autoencoder network architecture as exemplified here with three hidden layers. The equally sized encoder and decoder part share the bottleneck layer $\zv$. Tied weights are used indicated by the set of weights $(\Wv_1, \Wv_2)$ for the encoder and the transpose of these for the decoder part $(\Wv_2^T, \Wv_1^T)$. Note that bias weights are not explicitly shown in the figure. The output layer $\xt$ is the autoencoder reconstruction of the input $\xv$.}
\label{fig:model}
\end{figure}

The structure of the proposed autoencoder is based on stacking simple denoising autoencoders \cite{Vincent:2008} with some adjustments detailed in this section. The full autoencoder follows a butterfly construction with equal sized encoder and decoder parts, see Figure~\ref{fig:model}. Each layer in the encoder performs a deterministic mapping from layer input $\hv$ to layer output $\hv'$ with parameters $\Wv$ and $\bv$,
\begin{equation}
\label{eq:one_layer}
\hv' = f(\Wv\hv+\bv),
\end{equation}
where $f(\cdot)$ is the mapping (or activation) function. The weight matrix $\Wv$, and the bias vector $\bv$ are trainable parameters for each layer of the encoder. The same mapping process is then applied to the decoder part where the last layer of the decoder is the reconstructed input to the encoder. It is important to design the last layer of the encoder, denoted $\zv$, as a bottleneck to get a compressed representation of input and avoid identity mapping. 
The proposed autoencoder is also using tied weights, where the decoders weights are the transposed of the encoder weights. For instance if $\Wv'_L$ denotes the last weight matrix of the decoder, then $\Wv'_L = \Wv^T_1$, the transposed of the first weight matrix of the encoder (see Fig. \ref{fig:model}). This symmetric network lowers the probability of staying in the linear conformation \cite{Vincent}. Also, it is computationally faster, having tied weights as compared to independent weights for the decoder part. All hidden layers in the autoencoder use a $\tanh$ or Rectified linear unit (ReLU) activation function~\cite{relu}. The activation function for the output layer, which depends on the characteristics of the dataset, can be linear, logistic, softmax, or a combination of those.

\subsection{ Training the autoencoder}

In this study, we train the autoencoders in a two-step procedure. The first step consists pre-training of each layer of weights in the encoder part. After pre-training, all layers of the encoder are stacked and mirrored to form the decoder part (transpose of encoder weights), see Figure~\ref{fig:model}. The second step of the Training consists of fine-tuning of the full autoencoder. 

The pre-training step of the encoder follows closely that of Vincent et al. \cite{Vincent} by stacking simple (on the hidden layer) autoencoders trained with a denoising technique where a randomly chosen fraction of the inputs are set to zero for each sample (masking noise). The denoising is important to avoid identity mappings, especially if the size of the hidden layer is larger than the number of inputs. The amount of denoising used is a tunable hyper-parameter. Furthermore, we are using tied weights, also when training each simple autoencoder. The Training consists of minimizing the error between the input and the reconstructed output. If we assume a dataset with continuous values $\xv$, we use a summed square loss function when training each of the simple autoencoders, together with linear output activation functions. For the first layer, this is given by,
\begin{equation}
\label{eq:error_firstMSE}
E_1 = \frac{1}{N}\sum_{n=1}^N ||\xv_n - \xh_n||^2,
\end{equation}
where $\xh$ is the reconstructed input as given by the first simple autoencoder. 
\begin{equation}
\label{eq:xhat}
\xh = \Wv_1^Tf(\Wv_1\xv + \bv) + \bv'
\end{equation}

Pre-training of the second layer of the encoder consists of taking the first hidden layer representation of the data as input to a second simple autoencoder and minimizing the squared loss between the hidden representation and its reconstruction.
\begin{equation}
\label{eq:error_layer}
E_2 = \frac{1}{N}\sum_{n=1}^N ||\hv_n^{(1)} - \hh_n^{(1)}||^2
\end{equation}
Here the $\hh_n^{(1)}$ is the reconstructed first layer hidden representation given the second simple autoencoder. The remaining layers of the encoder are pre-trained by repeatedly training simple autoencoders with hidden layer representations as input data. Once all layers of the encoder have been pre-trained the decoder is automatically given by the tied weights method, see Figure \ref{fig:model}. Bias weights, however, are randomly initialized in the decoder part. 

Fine-tuning of the full autoencoder is accomplished by minimizing the loss function $E_f$,
\begin{equation}
\label{eq:error_full}
E_f = \frac{1}{N}\sum_{n=1}^N ||\xv_n - \xt_n||^2.
\end{equation}
Here $\xt_n$ denotes the full autoencoder reconstruction of the input $\xv_n$ (see Figure \ref{fig:model}). The minimization of $E_f$ is accomplished using stochastic gradient descent (with dense-matrix GPU implementation) \cite{stochastic}. In addition to improving the efficiency of stochastic gradient descent, three different methods to control the learning rate was available during training, Nesterov momentum, RMSProp, and Adam \cite{all_sgd}.  We used the dropout technique to avoid overfitting during the fine-tuning, both on the inputs and the hidden layers. We also allow for $L_2$ regularization during training, if necessary, to further reduce possible overfitting of the model. 

\subsection{Dealing with mixed data variables}
\begin{figure}
\centering
\includegraphics[scale=0.7]{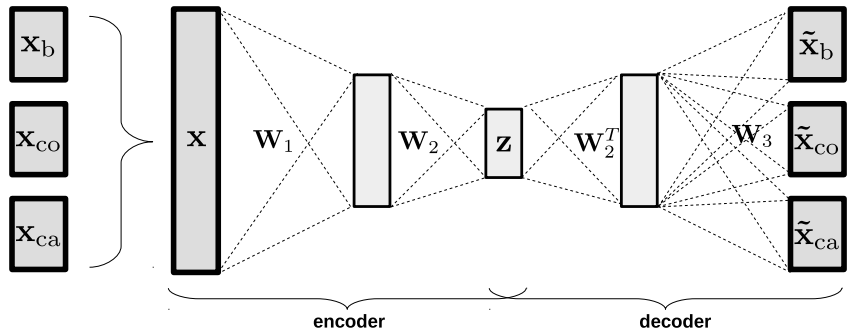}
\caption{The autoencoder architecture for data with a mixture of different variable types. The binary, continuous, and categorical inputs are concatenated to a single input vector but are separated in the output layer. Each variable type has a separate loss function. The autoencoder is not using tied weights in the last layer of the decoder when having more than one variable type.  
}
\label{fig:ae_mixed}
\end{figure}

Occasionally datasets consist of a mixture of different variable types, e.g., in medical data, it is widespread to have a combination of continuous, binary, and categorical variables. Pre-processing of binary or categorical variables often means using binary or one-hot encoding \cite{harris2010hot-coding}, respectively. The variable type affects the choice of the error function and output activation function when training the autoencoder, both in the pre-training and the fine-tuning step. For the case of binary data, the output activation is the sigmoid, and we use the cross-entropy loss function,
\begin{equation}
E_b = - \frac{1}{N}\sum_{n=1}^N\left[ \xv_{n}\log{\xt_{n}} + (1-\xv_{n})\log{(1-\xt_{n})}\right],
\end{equation}
where now $\xt_{n}$ denotes the prediction for sample $n$. The change of the reconstruction error also applies to the pre-training of the first layer of the encoder. All subsequent layers of the encoder use the squared error given by equation~\ref{eq:error_layer} regardless of the input data type. For the case of mixed continuous, binary and categorical inputs the reconstruction error is a sum of both squared and cross-entropy loss (see Figure~\ref{fig:ae_mixed}), with the appropriate choice of linear, sigmoidal and softmax output activation functions. As illustrated in Figure~\ref{fig:ae_mixed} the mixed input data is concatenated to form a single input vector to the encoder. To ease the implementation, the last layer in the decoder does not use the transpose of the first layer of the encoder. 

\subsection{Autoencoder for imputation}\label{sec:sdai}
To be able to use the autoencoder as a tool for imputation of missing data, two additional steps are needed. First, the autoencoder must be able to train on a dataset that contains missing data. Second, given a trained autoencoder, how can this trained autoencoder reconstruct an incomplete input? 

For the first step, missing data for each variable are replaced by the average over the known ones. The loss function used during training is affected by missing values; therefore, all components of the input vector $\xv_n$ that contain missing values are masked out when forming the error. For instance, in the case of continuous input data, the mean squared loss function is calculated over known values only. Equation~\ref{eq:error_firstMSE}, showing the loss function for the first layer, is therefore modified according to,  

\begin{equation}
\label{eq:error_firstMSE_mask}
E_1 = \frac{1}{N}\sum_{n=1}^N ||\xv_n^{\text{(known)}} - \xh_n^{\text{(known)}}||^2.
\end{equation}

Masking of missing values when computing the reconstruction error is also used for binary and categorical input data and applies to both pre-training and fine-tuning of the autoencoder.

With the trained autoencoder, it is straightforward to impute missing values of an input vector. Similar to the training step, all missing values are replaced by mean values when entered into the encoder. The corresponding output of the decoder then gives the imputed values. For continuous data, we use the actual decoder outputs, and for binary or categorical data, the outputs can be interpreted as class probabilities or be converted into exact class predictions. 
We denote the proposed autoencoder for imputation of missing data by SDAi (Stacked Denoising Autoencoder for Imputation).

\section{Datasets}\label{sec:datasets}

Our model for imputation of missing data (SDAi) was tested on six different datasets with different characteristics, including one dataset with a mixture of both numerical and categorical input variables. Four of the datasets were synthetic and generated from the same underlying formula (see below). Table \ref{table:all-data} shows a summary of the datasets used in this study.

\begin{table}[H]
\centering
\caption{Sample sizes and number of features for all dataset used in the numerical experiments.}
\begin{tabular}{|c|c|}
\hline
data & (samples, features) \\
\hline
Synthetic, $n_f = 1$ & $(1000,200)$\\
Synthetic, $n_f = 4$ & $(1000,200)$\\
Synthetic, $n_f = 2$ & $(2000,500)$\\
Synthetic, $n_f = 4$ & $(2000,500)$\\
MNIST            & $(70000, 784) $\\
Proteomic        & $(945, 134)$ \\
Genomic          &  $(3738, 978)$ \\
NYC bicycle      & $(856, 6)$ \\
Olivetti faces   & $(400, 4096)$  \\
\hline
\end{tabular}%
\label{table:all-data}
\end{table}

\subsection{Synthetic data}\label{ss:synth_data}
We generated $d$-dimensional data by shifting sine functions with known frequencies (one or four random frequencies are used in the actual runs) and a random phase for each sample. Independent Gaussian noise is added to each feature in the samples to make the problem more challenging. The synthetic dataset $D_{Nd}$ with $N$ samples and $d$ features is generated as follows: 
\begin{align*}
D_{Nd}&=\{\sv_1,\ldots,\sv_N\},\quad \textmd{where} \quad \sv_n\in \mathop{{}\mathbb{R}^d}, \quad n=1,...,N.\\
s_{i,n}&=\sin(f_n x_i)+z_i, \quad x_i=2\pi\frac{i-1}{d}+u_n, \quad \forall i=1,\ldots,d,\\
\textmd{where}&\quad u_n \sim \mathcal{U}(0,2\pi) \quad \textmd{and} \quad
z_i \sim \mathcal{N}(0,\sigma^2). \quad 
\end{align*}
In the experiment, we generated two different sizes of this data, $1000$ and $2000$ samples with $200$ and $500$ variables respectively, with different random frequencies. The frequency $f_n$ for each sample is randomly chosen from a predefined set of frequencies (integers) for each data set. Using more than one frequency adds more complexity to the data.

\subsection{MNIST data}
The algorithm was also tested on an image dataset of handwritten digits, the MNIST data \cite{mnist}. The data contains $70,000$ images, where each image is of size $28\times 28$ pixels. We used $50,000$ for training and $10,000$ for validation and testing each. The images were normalized such that each pixel $\in [0,1]$. 
\begin{figure}
\centering
\includegraphics[scale=0.3]{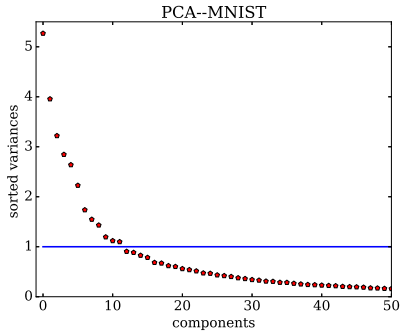}
\caption{Shows sorted explained variances, which have been calculated by projecting the data using PCA. The horizontal blue line shows the Kaiser criterion}
\label{fig:pca_mnist}
\end{figure}
A PCA analysis on the MNIST data reveals that only a handful of directions carry most of the variation (see Figure \ref{fig:pca_mnist}). The blue line shows the Kaiser criterion~\cite{jackson}, which separates the components of data with eigenvalues above one from the rest. This criterion can give a perspective of the range of the bottleneck's size for the network architecture.

\subsection{Gene and protein expression data}\label{sec:bio-data}
One proteomic dataset \cite{arrayexpress} (protein profiling of breast cancer cell lines), consisting of $134$ variables (antibody expression measurements) in $945$ samples, was used. For this data, the Kaiser criterion shows that $22$ PCA components carry most of the variation. 

An additional biological dataset, gene expression measurement for somatic mutations in lung cancer \cite{arrayexpress} was used. It contains $3738$ samples with gene expression of $978$ transcripts. As for the proteomic dataset, there is a large degree of linear correlation among the $978$ variables, as indicated by the PCA analysis.

\subsection{Bicycle counts for east river bridges}
This dataset contains a mixture of continuous, binary, and categorical variables. The data has been collected during seven months in the year $2016$ on four east river bridges in New York city\footnote{NYC Open Data \url{https://data.cityofnewyork.us/Transportation/Bicycle-Counts-for-East-River-Bridges/gua4-p9wg}}. The dataset originally contains $214$ samples and $10$ features. 

We reshaped the structure of the data to increase the number of samples and add categorical variables. Each sample now contains only one bridge and where the bridge name is added as a categorical variable. Furthermore, the variable \textit{number of the bikes} on each bridge, was added to the data, and the precipitation variable was transformed to a binary variable \textit{rain}, which means any positive precipitation is one, and the rest is zeros. Finally, the only considered temperature features were the average of the low and high temperature of each day. After these changes, we ended up with $856$ samples and six features:
\begin{itemize}[noitemsep]
\item Bridge name (Categorical)
\item weekday (Categorical)
\item Average Temperature (Continuous)
\item Nof bikes on each bridge (Continuous)
\item Total number of bikes on each day (Continuous)
\item Rain (Binary).
\end{itemize}

\subsection{The Olivetti faces dataset}
\begin{figure}
\centering
\includegraphics[scale=.75]{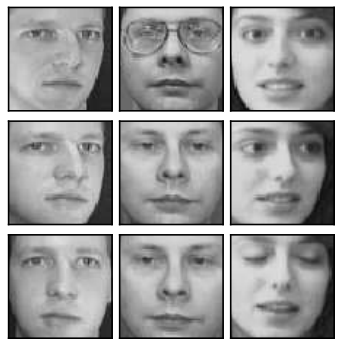}
\caption{A random selection of nine images from the Olivetti face dataset provided by AT\&T Laboratories Cambridge. Each column shows samples from the same face.}
\label{fig:oliv}
\end{figure}

AT\&T Laboratories Cambridge provides $400$ images of faces\footnote{The Database of Faces \url{https://www.cl.cam.ac.uk/research/dtg/attarchive/facedatabase.html}}, each of size $64 \times 64$. There are in total $40$ subjects in $10$ different situations, e.g., varying lighting conditions, facial expressions, etc. Figure~\ref{fig:oliv} shows a subset of nine images, three samples for three different faces. Before using the face data in the numerical experiments they were normalized such that each pixel $\in [0,1]$.

\section{Numerical experiments procedure}\label{sec:experiments}
The datasets were used to test the proposed SDAi imputation method. For comparison, four other popular methods were also tested on the same datasets. The first one was an improved K-nearest neighbor (K-NN) imputation method by Tutz et al.~\cite{knn}. We also used the well known multiple imputation technique MICE~\cite{mice} as the second external method in the experiments. As the third method we considered a random forest imputation (RF) model \cite{missforest} based on regression and classification trees. They represent an ensemble learning method that can handle continuous and categorical data. Finally, as a baseline method we used mean imputation.  
\subsection{Hyper-parameters}\label{sec:hyper}
The following hyper-parameters for the SDAi model were optimized during model selection:
\begin{itemize}[noitemsep]
\item Number of hidden layers
\item Hidden layer sizes
\item Regularization parameters
\item Dropout probabilities
\item Mini-batch size
\item Activation function
\item Learning rates
\item Number of epochs
\end{itemize}

For the Olivetti faces dataset we have used autoencoder with convolutional layers~\cite{lecun1999cnn} to allow for a more efficient architecture. Here the structure did not follow a butterfly form, and for the decoder only one deconvolutional layer was used. Additional hyper-parameters for the convolutional layers were:
\begin{itemize}[noitemsep]
    \item Number of output filters
    \item Kernel size
    \item Strides 
    \item Padding
\end{itemize}

Because of the high dimension of hyper-parameter space, random search \cite{hyper_random} was used to select the optimized model, although some heuristics were used to find initial values of some of the parameters. On each parameter, an interval of possibilities was defined, which varied with the datasets. For instance, the possible range for the bottleneck was guided by a PCA analysis. After selection of the initial hyper-parameter sets, a random search among all of the shuffled series was performed. The error for each model was estimated using 5-fold cross-validation (see next section), and the random search was terminated after a fixed number of trails after which the best performing model was selected.

As for K-NN~\cite{knn} and MICE~\cite{mice}, they both use different hyper-parameters. In this study, we tuned two hyper-parameters present in MICE, the number of multiple imputations and the number of iterations in each imputation. Both of these parameters were determined based on a convergence criterion of the model parameters. For the NYC bicycle data, the data with both numerical and categorical variables, we used Bayesian linear regression for numeric, logistic regression for binary and multinomial logistic regression for multiclass variables. 
In the K-NN model selection, the distance weight regularization parameter, $\lambda$~\cite{knn}, and the number of neighbors (K) were tuned. For the RF imputation method we only tuned the number of iterations of the imputation process and the number of trees used in the ensemble.

\subsection{The general setup}\label{sec:expsetup}
All datasets used in the experiments were complete, i.e., they contained no missing values. A random part of the data was therefore removed (corrupted) and used as the gold standard when evaluating the different methods to impute the removed portion. For all non-image datasets we used the root mean squared error (RMSE) between the imputed data and the gold standard to evaluate the performance of the different methods. Since the NYC bicycle data contains a mixture of numerical and categorical, we used the sum of the RMSE (for the continuous part) and the cross-entropy (for the categorical part) for the evaluation of the methods.
Different amounts of corruption were tested, ranging from $10\%$ to $60\%$ for almost all datasets. For the MNIST data up to $90\%$ corruption was tested, however for the Olivetti faces only a single instance of local masking was used, resulted in approximately $65\%$ corruption. The corruption process was done before the splitting of data into training and test, meaning that the same amount of missing data was also used during the training of the SDAi method. For the K-NN, MICE and RF, the hyper-parameter selection and imputation were accomplished using the entire data. This procedure was repeated for each fraction of corruption. 

For SDAi, to avoid biases from possible overfitting, a two-loop testing and model selection strategy was employed (see Figure \ref{fig:alg}). The innermost loop is for model selection, where hyper-parameters are determined given a training dataset. The resulting optimal model was then tested on data from the outer test loop, not part of any training or model selection. Due to long computational times, we used a single fixed validation and test set for the MNIST data. 

All five imputation methods were evaluated on precisely the same test sets, as defined by the outer cross-validation loop.
\begin{figure}[H]
\begin{algorithm}[H]
\SetAlgoLined
\For{fraction of corruption $\in \{10\%,\dots,60\%\}$}{
    $5$-fold cross-validation split\;
    \For{test $\in$ test sets}{
        $5$-fold cross-validation split\;
        \For { hyper-parameters $\in$  distributions}{
        \For{train $\in$ training sets}{
            train and validate the models with randomly selected hyper-parameters\;}
        compute the average error over all folds\;}
        select optimal hyper-parameters corresponding to the best average performance\;
        train a new model with optimal hyper-parameters on the full training data\;
        calculate the performance on the test set\;}
     average and standard deviation of all test set performances\;}
\end{algorithm}
\caption{The training and test process used when evaluating the SDAi method. Two loops were used, the outermost loop was for estimation of test performance and the innermost loop for model selection, given a specific training dataset. The whole procedure was repeated for different fractions of initial corruption.}
\label{fig:alg}
\end{figure}
\section{Results}\label{sec:results}
\subsection{SDAi hyper-parameters}
The optimal set of hyper-parameters for the SDAi method, for a given problem, varies with the amount of corruption of the datasets. Typical values for $30\%$ corruption for all datasets that are achieved with the random search are shown in Table~\ref{table:hyper1}~\&~\ref{table:hyper2}. The first four rows show the values for different realizations of the synthetic dataset. A general trend can be found regarding the use of dropout training, layers close to the bottleneck generally have no or little dropout (small probability of removing nodes).  
\begin{table}
\centering
\caption{The table shows a set of hyper-parameters for the SDAi method obtained during model selection for $30\%$ of manually corrupted data. Network size refers to the encoder part of the model (layer sizes), where the last number corresponds to the size of the latent space. For the regularization, $\lambda$ refers to the parameter controlling the scale of $L2$ and dropout being the probability of removing nodes during training.}
\begin{tabular}{|| c ||c || c || c ||} 
\hline
 Dataset & Network size & Dropout &  $\lambda\;(L2)$\\\hline 
  Synth (1000,200) $n_f=1$ & $[100,8]$&  $[0.3, 0.1]$ & $10^{-4}$   \\ \hline
  Synth (1000,200) $n_f=4$ & $[100,50,12]$&  $[0.3, 0.1, 0]$ & $10^{-3}$ \\ \hline
  Synth (2000,500) $n_f=1$ & $[160,50,15]$&  $[0.4, 0.2, 0.1]$ & $10^{-4}$  \\ \hline
  Synth (2000,500) $n_f=4$ & $[250,100,15]$&  $[0.3, 0.2, 0.1]$  & $10^{-4}$  \\\hline
  MNIST 				  & $[209, 111, 55]$& $[0.2, 0.1, 0]$  &  $10^{-5}$  \\ \hline
  Proteomic data          & $[100,30]$ & $[0.3, 0.1]$ & $10^{-3}$   \\ \hline
  Genomic data            & $[300,50]$ &  $[0.3, 0.2]$ & $10^{-4}$     \\ \hline
   NYC bicycle data& $[29, 5]$ &  $[0.1, 0]$ & $10^{-2}$     \\ \hline

\end{tabular}\label{table:hyper1}

\vspace{1cm}
\caption{The selected convolutional hyper-parameters for Olivetti face data. Only one convolutional layer for the encoder was found to be optimal.}
\begin{tabular}{|| c ||c || c || c ||c ||c||} 
\hline
 Dataset & Filter size & Kernel size & Latent dimension & Strides\\\hline 
  Olivetti dataset & $[77]$&  $[10]$ & $90$ & $2$  \\ \hline
  \end{tabular}\label{table:hyper2}

\end{table}

\subsection{Synthetic data}\label{sec:synthetic}
\begin{figure}[H]
    \centering
    	\begin{minipage}{5cm}
    		\centering
    		\includegraphics[scale=0.4]{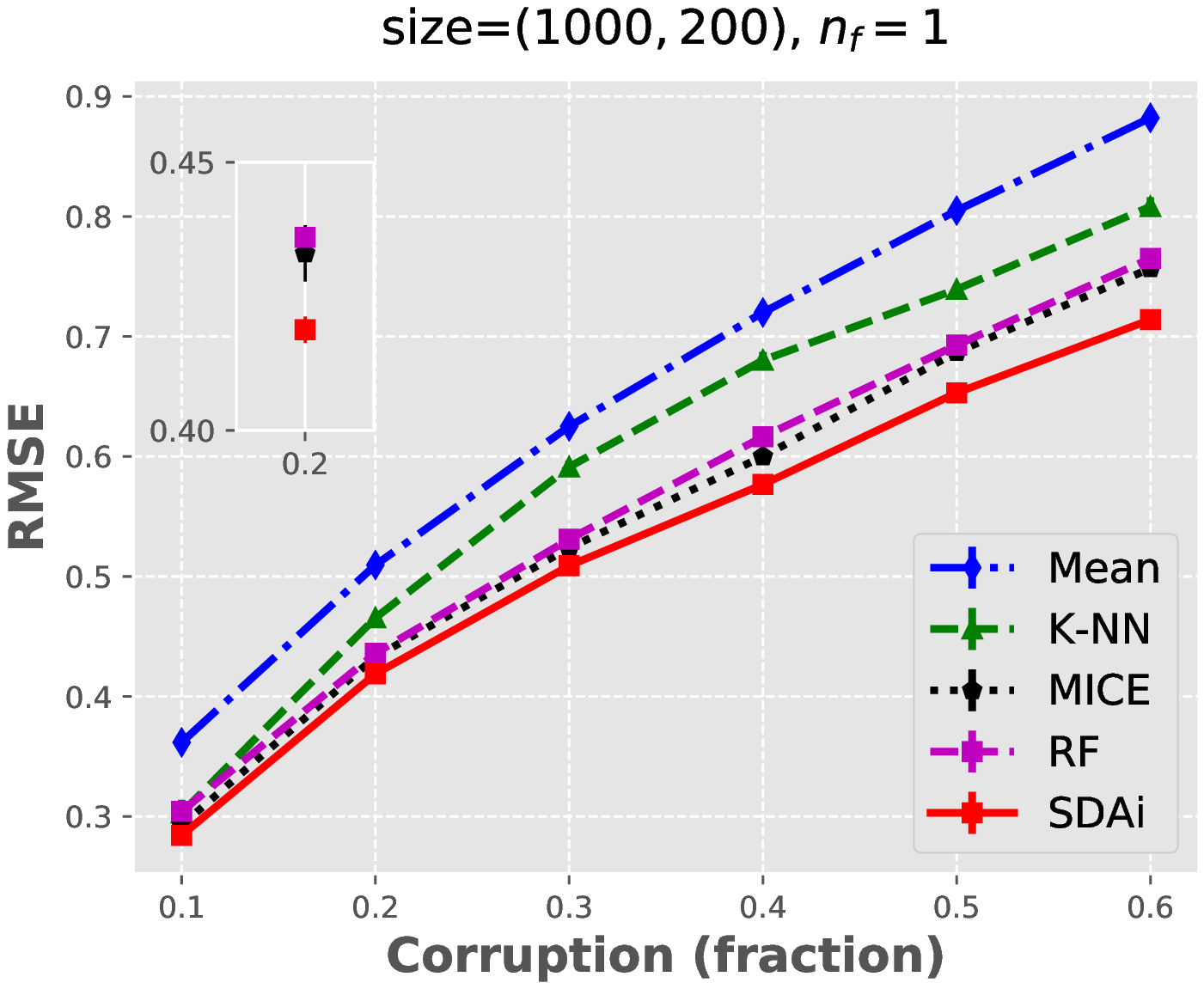}
    			\includegraphics[scale=0.4]{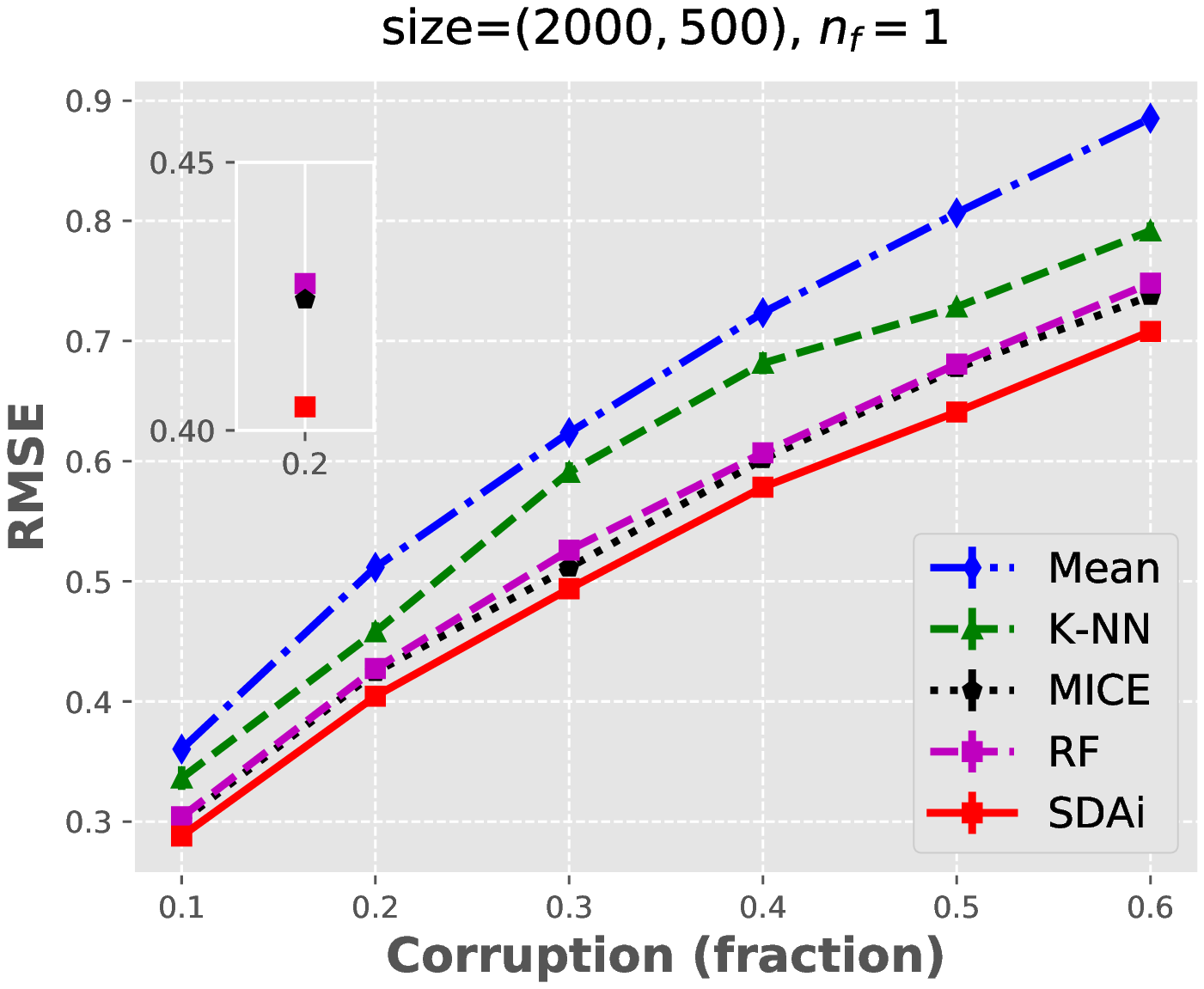} 
    	\end{minipage}
    	\hspace{2cm}
    	\begin{minipage}{5cm}
    		\centering
    		\includegraphics[scale=0.4]{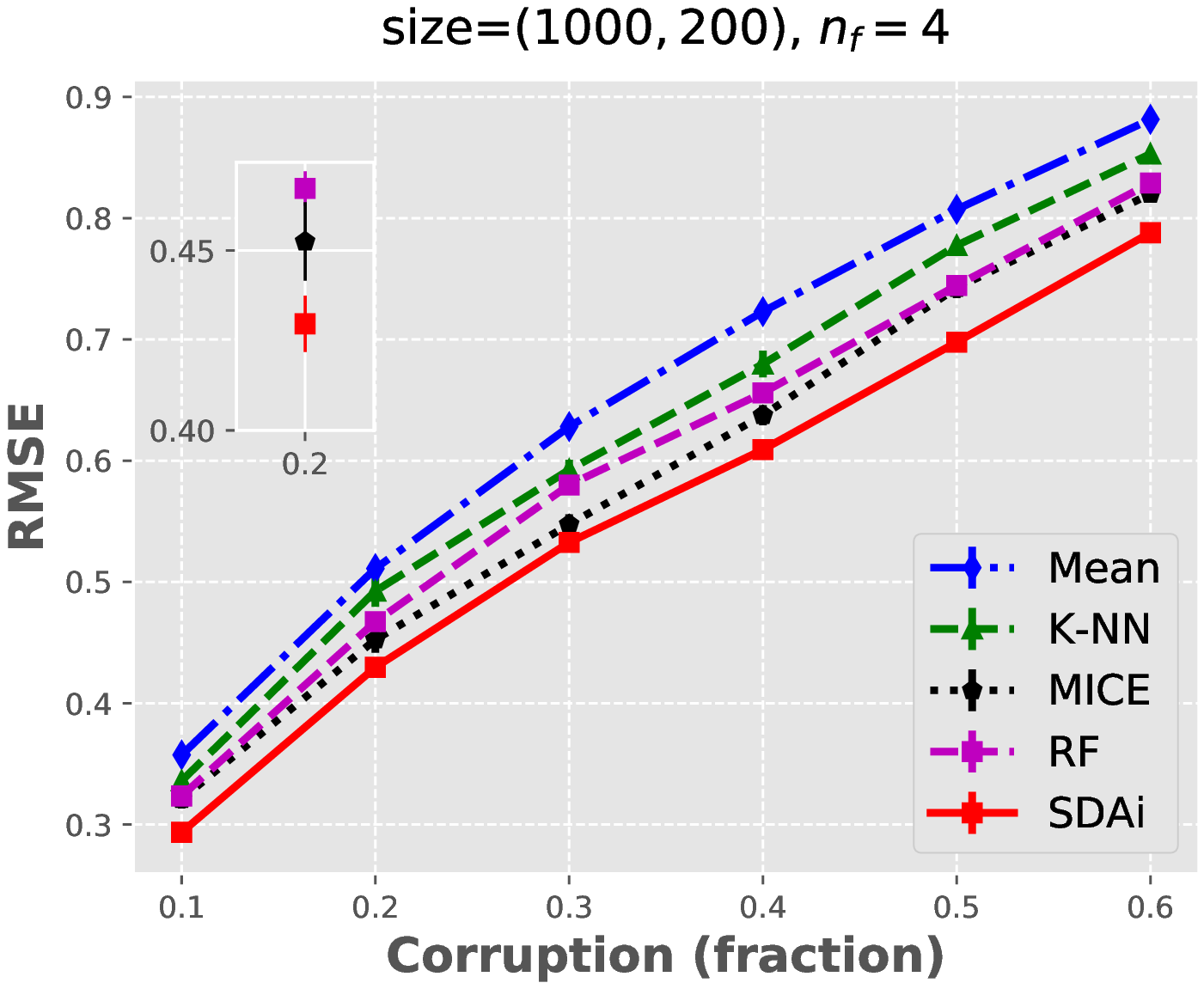}
    		\includegraphics[scale=0.4]{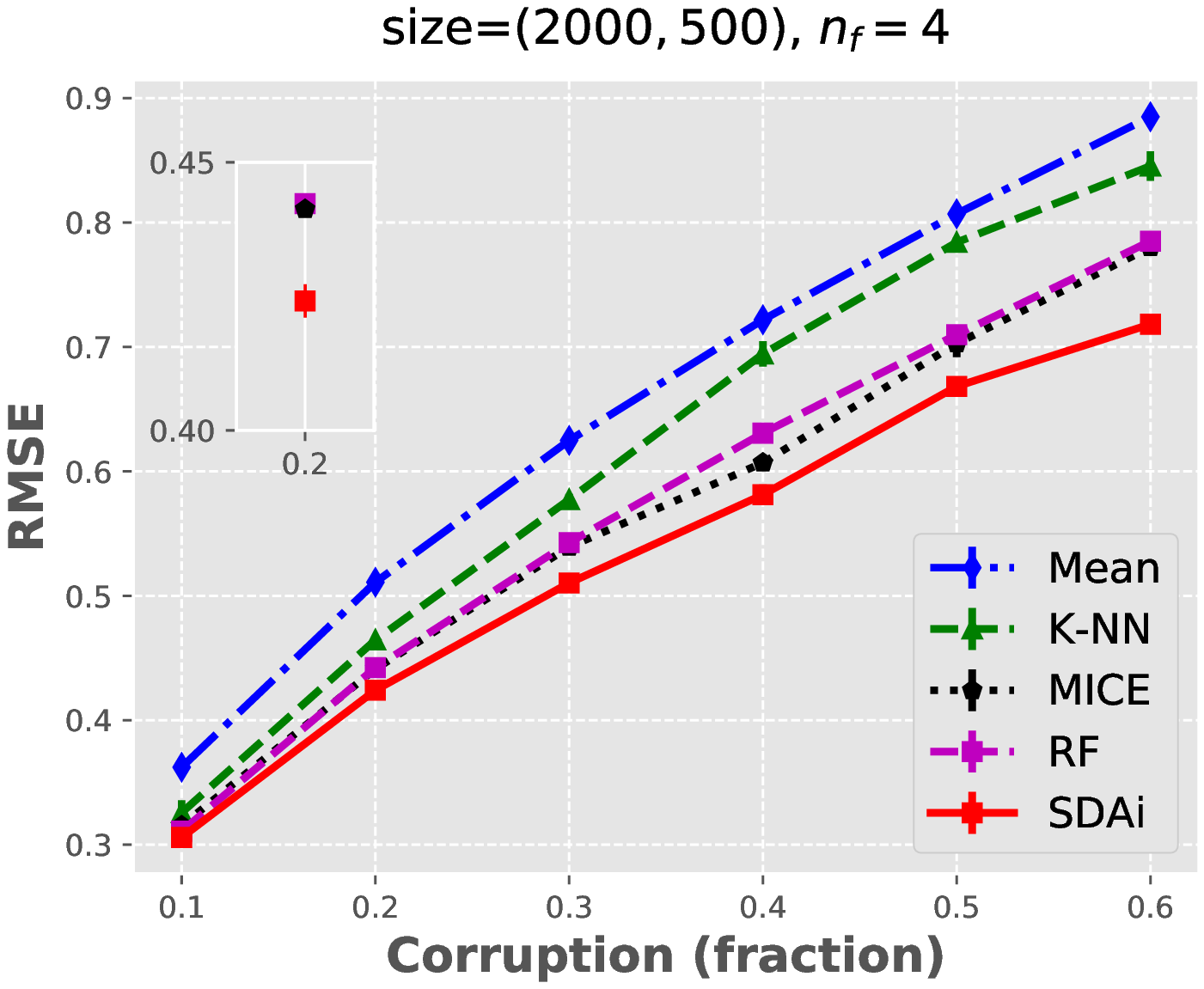}
    	\end{minipage}
  	  	\caption{Root mean squared (RMSE) imputation error for the test data as a function of the amount of corruption during training. Top left: Synthetic data with $1000$ samples, $200$ features, and one frequency component. Top right: The same size as the top left, but with four frequency components (more complexity). Down left: Larger data size with $2000$ samples and $500$ features (one frequency). Downright: The same size as down left, but with four frequency components (more complexity). The standard deviations (error bars) across the different test sets are small and hardly visible. A zoom-in is shown for $20\%$ corruption where the size of the error bars can be seen.}
\label{fig:syn}
    \end{figure}

We used four different realizations of the synthetic dataset with varying sizes and degree of complexity, as detailed in section \ref{ss:synth_data}. Two datasets with 1000 samples and 200 features, and two larger datasets with 2000 samples and 500 features where generated. In each case, the two datasets were different corresponding to the number of frequencies ($n_f$) used. We used $n_f=1$ and $n_f=4$ where the latter represents a dataset with more complexity as compared to the former. Figure \ref{fig:syn} shows the root mean squared error (RMSE) for the imputed test data as a function of the degree of initial random corruption of the data. Imputation using simple mean values consistently resulted in the largest RMSE. For all four synthetic datasets, SDAi method was found to give the lowest RSME for all ranges of corruption. After comparing the size of selected autoencoder architectures by random search, larger and more complex datasets require more extensive and more layers of the SDAi model.

\subsection{MNIST}\label{sec:mnist}
\begin{figure}
\centering
\includegraphics[scale=0.7]{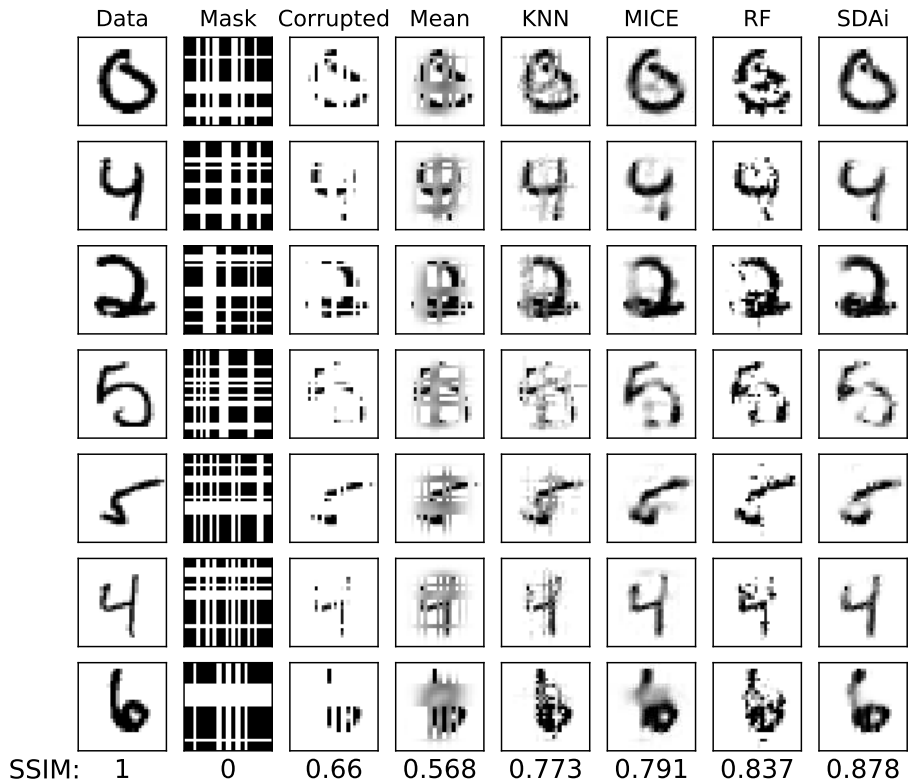} 
\caption{The first column from the left shows seven samples of the original test data. In the third column, $60\%$ of the data has been corrupted by lines in the second column. The last five column shows the imputed images using Mean, K-NN, MICE, RF, and SDAi, respectively.}
\label{fig:mn6}
\end{figure}

\begin{figure}
\centering
\includegraphics[scale=0.6]{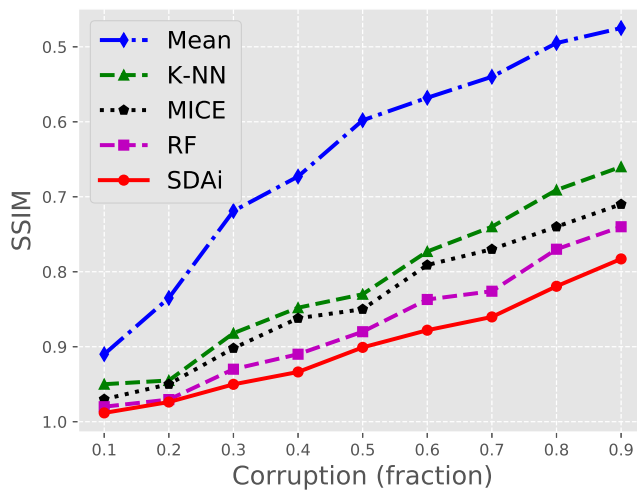}
\caption{SSIM for the MNIST test data as a function of the amount for corruption during training. To have a similar appearance as for previous plots, the y-axis is reversed, starting from SSIM $= 1.0$ and decreasing.}
\label{fig:mn}
\end{figure}

For the MNIST image dataset, we used a single validation and test dataset of $10,000$ images each which left $50,000$ images for the training. To challenge the imputation methods, and because of the high correlation of the image data, the corruption process of the images acted on lines of the images rather than on individual pixels. For example, an approximately $50\%$ corrupted image means that a random selection (without replacement) of $25\%$ of the horizontal and $25\%$ of the vertical lines were missing. Figure~\ref{fig:mn6} shows a selection of seven images from the test set with the original, mask, corrupted, and the results of imputation methods images. The $60\%$ corrupted version is the input to the different imputation methods, and the resulting imputed images are shown in the next columns. Visually the SDAi methods were able to recover the original images in a better way compared to MICE, RF, K-NN, and Mean imputation.
To analyze the outcomes to the original images instead of using RMSE, we used the structural similarity index (SSIM)\cite{wang-SSIM:2004}. With RMSE, the cost is calculated on the distance between the pixels, that does not necessary shows the loss of quality in images (images with equal RMSE might not contain the same noises), whereas SSIM depends on structural features which are extracted from the image.
Figure \ref{fig:mn} shows the SSIM for a varying degree of corruptions (smudged random horizontal and vertical lines), up to as much as $90\%$. Again the SDAi method obtains the smallest SSIM over all ranges of corruption.
\subsection{Gene and protein expression data}\label{sec:bio}
\begin{figure}[H]
\centering
    	\begin{minipage}{6cm}
    		\centering
    		\includegraphics[scale=0.4]{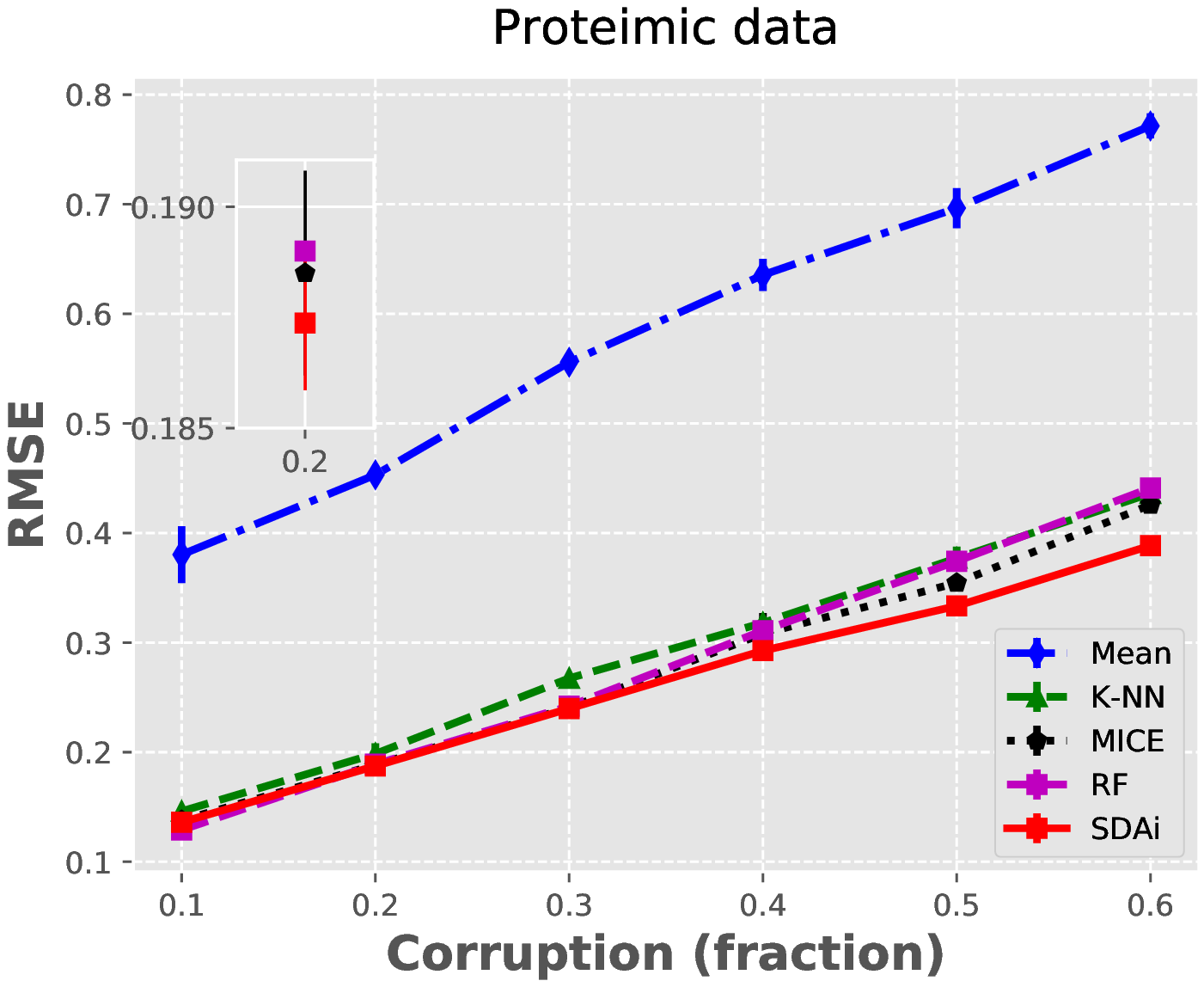}
    	\end{minipage}
\hspace{1cm}
    	\begin{minipage}{6cm}
    		\centering
    		\includegraphics[scale=0.4]{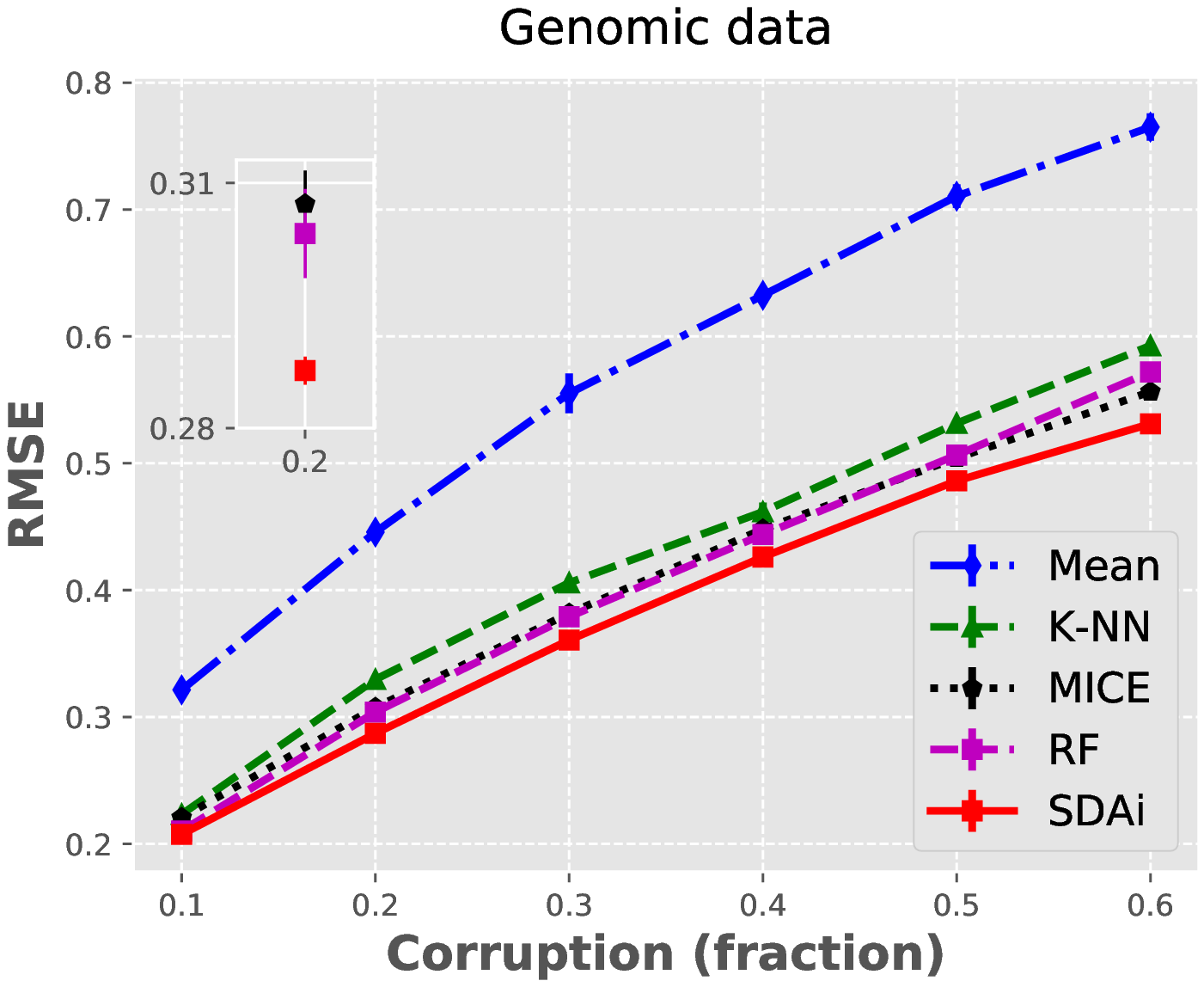}
    	\end{minipage}
\caption{Root mean squared (RMSE) imputation error for the test data as a function of the amount of corruption during training. Left: (proteomic data) After $40\%$ of corruption, SDAi shows an improvement of estimation compare to MICE, RF, and K-NN. Right: (genomic data) SDAi shows lower RMSE compare to the other methods.}
\label{fig:proteomic}
\end{figure}

Due to the small size of the proteomic data, a $4$-fold cross validation was used. Figure \ref{fig:proteomic} shows the RMSE for varying degrees of corruption. We can conclude that for the proteomic dataset, K-NN, MICE, RF and SDAi showed similar results. For the genomic dataset, the SDAi method obtained the best imputation results.

The relatively small bottleneck size for the proteomic data is probably due to the high correlation of the variables. The genomic data on the other hand, required a larger bottleneck size indicating less correlation among the input variables. 
\subsection{New York bicycle counts data}
\begin{figure}
\centering
\includegraphics[scale=0.5]{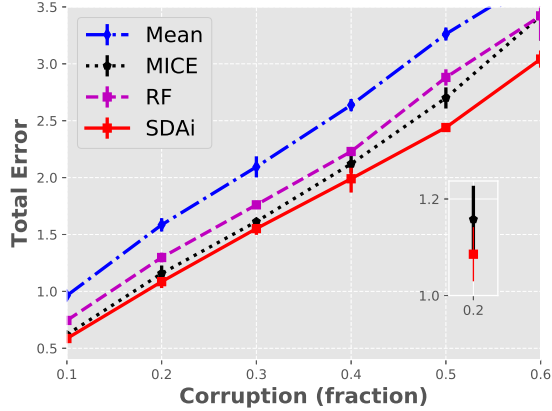}
\caption{The sum of all the errors for the different type of variables are presented as \textbf{total error} in the y-axis. SDAi shows better results in all different portions of corruption.}
\label{fig:bicycle}
\end{figure}

For bicycle data, which contains a mixture of different variables, again SDAi improved the imputation results compared to the other methods. The error in figure~\ref{fig:bicycle} is the sum over logistic loss and RMSE for all the variables presented in the data. For the MICE method, we used Bayesian linear regression, logistic regression, and multinomial logistic regression models for numeric, binary, and categorical variables, respectively. 
The categorical variables in this data have no natural orders; therefore, we encode them to one-hot vectors to have a suitable numeric form of variables for statistical processing. Since K-NN by definition uses Euclidean distance, it is possible to use it on all variables. However, the results show that since the one-hot encoded vectors have the same length, K-NN does not find the higher correlation in this data. As the K-NN imputation error was more elevated than Mean imputation error, we omitted K-NN in figure \ref{fig:bicycle}.

\subsection{The Olivetti face dataset}
\begin{figure}
\centering
\includegraphics[scale=.77]{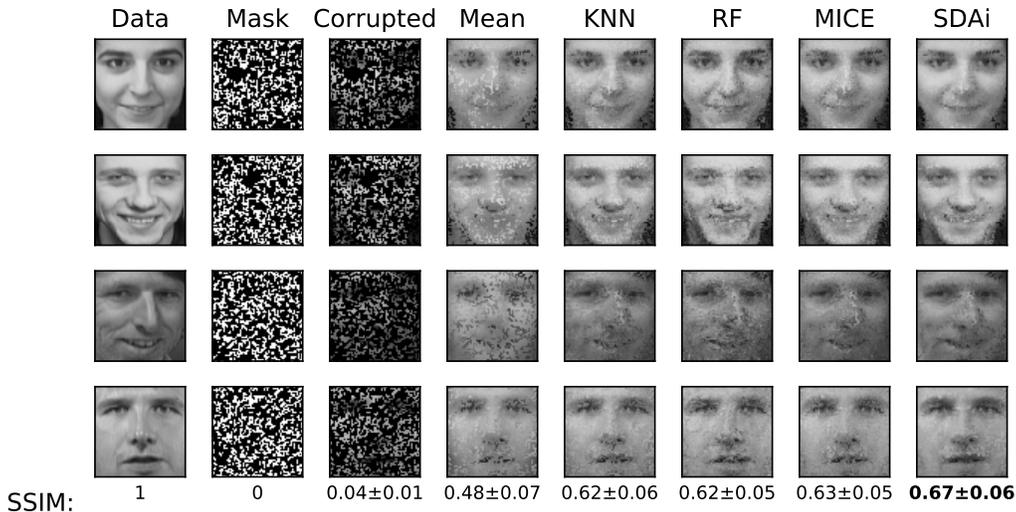}
\caption{The first column from the left shows four randomly selected faces from the original test data. In the third column, approximately $65\%$ of the data has been corrupted by Ising-like mask in the second column. The last five columns show the imputed images using Mean, K-NN, RF, MICE, and SDAi, respectively. The last row shows averaged SSIM with standard deviation for all imputed images in the test data.}
\label{fig:face}
\end{figure}

Due to the high local correlation in the face data, the line corruption used for the MNIST data is not challenging enough. Instead, we used a Square-lattice Ising-like~\cite{ising} corruption mask, $64\times64$ in dimension, resulting in approximately $65\%$ of the pixels corruption, i.e., replaced by NaN.  The first three columns in Figure~\ref{fig:face} shows original faces, the mask, and the corrupted face after applying the mask, respectively. The remaining columns shows the imputed images using the different methods. Except for the mean imputation method they all, visually, appear to give similar imputed images. However there is an advantage for the SDAi method when comparing SSIM values (last row in Figure~\ref{fig:face}). In fact a Wilcoxon signed-rank test showed a significant improvement for the SDAi method ($p < 0.001$) against MICE.  

\subsection{Computational times}
Computational requirements for the different methods were estimated at 30\% corruption level for all datasets except for the Olivetti face data that used a fixed corruption at 65\%. The running times (in seconds) are presented in Table~\ref{table:time}. For the SDAi method, the running times included both training and test requirements and was measured using both the CPUs and a GPU implementation. The model selection procedure for all methods are not included in the timing report. The time for all methods have been measured on four logical CPUs, and for the GPU version of SDAi we used one NVIDIA card (GeForce GTX 1080 Ti).
\begin{table}[H]
\caption{Estimates of computational times (in seconds) at $30\%$ corruption for all datasets, except for Olivetti faces which used at a fixed percentage of corruption. For the SDAi method, the times include both training and test procedures.}
\resizebox{\columnwidth}{!}{%
\begin{tabular}{c|cccccc}
data$\backslash$ method & Mean & K-NN & Mice & RF & SDAi (CPU) &SDAi (GPU)  \\
\hline
Synth (1000,200) & 1 & 6 & 210 & 250 & 5 & 2\\
Synth (2000,500) & 1 & 14 & 670 & 840 & 5 & 2\\
MNIST            & 1 & 360 & 32,800 &  28,900 & 95 & 39 \\
Proteomic        & 1 & 1 & 170 & 250 & 23 & 6\\
Genomic          & 1 & 110 & 1960 & 9100 & 38 & 20 \\
NYC bicycle      & 1 & - & 180 & 3 & 6 & 7 \\
Olivetti faces   & 1 & 10 & 420 & 9700  & 470 & 22\\
\end{tabular}%
}
\label{table:time}
\end{table}

\section{Discussion and conclusions}\label{sec:discussion}
In this study, we have compared a general and flexible method for imputation of missing data, based on stacked denoising autoencoders, with some well-known techniques. The explorations showed consistently smaller imputation error for our proposed SDAi method compared to the other techniques (MICE, K-NN, and RF) across all tested datasets. All experiments were carried out using a two-loop testing and validation procedure to avoid overfitting and to obtain reliable test results. Furthermore, all comparisons were made with the assumption that complete data was not available during the training of the methods.

To entirely make use of data with a significant fraction of missing values, it is essential to impute the missing data reliably. Often data can be expensive and difficult to obtain, so discarding cases with missing data can result in too small sample sizes and may introduce biases. It is therefore essential to use a valid and accurate imputation method. We believe that the autoencoder, here implemented as imputation method (SDAi), is such a method. In all of the numerical experiments, we used the same amount of initial missing data for both training and testing to simulate a possible realistic scenario. As expected, the Mean imputation method gave the most substantial error. We also found MICE (using the average of many multiple imputations) and RF to be better than the K-NN method. However, for all ranges of the initial fraction of missing data, we found the autoencoder (SDAi) to consistently impute missing data with the smallest error, especially for datasets with a large amount of initial missing data.

The SDAi method is based on the pre-training of each layer of the encoder. While pre-training is known to prevent overfitting \cite{deep_learning} and combat the vanishing gradient problem, it can mostly be replaced by efficient regularization techniques and the use of the rectifier activation function. Nevertheless, we found pre-training always to be advantageous (smaller imputation error). This finding was based on the comparison (results not shown) where pre-training was replaced by standard random initialization of all encoder weights, followed by the fine-tuning session. We found for the small datasets that pre-training can reduce optimization time and more easily prevent overfitting. For big datasets, the difference between with and without pre-training estimations was less visible.

One of the datasets in this study, NYC bicycle data count, contains a mixture of continuous, binary, and categorical variables. There are two categorical (nominal) variables that have no intrinsic ordering. The typical approach is to encode these variables to one-hot vectors. With the separated input and output of the autoencoder, we were able to calculate the performance separately on each type of data. The results showed that SDAi could make a better imputation compared to MICE and RF. 

A challenge for the SDAi method and neural networks is model selection. There are more hyper-parameters to consider compared to \textit{e.g.} MICE, RF and K-NN. Here we found random search \cite{hyper_random} to be effective. We tested many combinations of hyper-parameters, and mostly after a few runs, it was possible to narrow down the range of promising hyper-parameter values. To initialize the scale of these sets,  we used unsupervised techniques such as PCA. Furthermore,  to overcome the optimization of the SDAi model, we used GPU computation as much as possible. However, the purpose of this study was not to reduce computation time, and no systematic effort was made to accomplish this.

In conclusion, we have found the proposed autoencoder framework to be very useful in missing data imputation, even in situations where there is a large degree of initial missing data. A natural next step is to extend the SDAi method to allow for multiple imputation using variational autoencoders.

\section{Acknowledgments}
We thank Professor Carsten Peterson for valuable discussions. The study was supported by funds from the Swedish Foundation for Strategic Research (CREATE Health).

\bibliographystyle{unsrt}  
\bibliography{sdai}

\end{document}